\documentclass{article}
\usepackage[verbose=true,letterpaper]{geometry}
\usepackage[numbers,compress]{natbib}
\usepackage[utf8]{inputenc} 

\makeatletter
\@namedef{opt@inputenc.sty}{utf8}
\makeatother
\usepackage{CJKutf8}

\usepackage[T2A,T1]{fontenc}    
\usepackage[russian,english]{babel}

\usepackage{url}            
\usepackage{booktabs}       
\usepackage{amsfonts}       
\usepackage{nicefrac}       
\usepackage{microtype}      
\usepackage{xcolor}         
\usepackage{graphicx}
\usepackage{amsmath}
\usepackage{amssymb}
\usepackage{multirow}
\usepackage{enumitem}
\usepackage{algorithm}
\usepackage{algorithmic}
\usepackage{colortbl}
\usepackage{array}
\usepackage{comment}
\usepackage{float}
\usepackage{authblk}
\usepackage{hyperref}

\title{Training-Free Tokenizer Transplantation\\
       via Orthogonal Matching Pursuit}
\author[]{Charles Goddard}
\author[]{Fernando Fernandes Neto}
\affil[]{\texttt{\{charles,fernando\}@arcee.ai}}
\date{}

\newcommand{\zhchar}[1]{{\begin{CJK*}{UTF8}{gbsn}#1\end{CJK*}}}
\newcommand{\kochar}[1]{{\begin{CJK*}{UTF8}{mj}#1\end{CJK*}}}
\newcommand{\jachar}[1]{{\begin{CJK*}{UTF8}{gbsn}#1\end{CJK*}}}

\definecolor{posstrong}{rgb}{0.0, 0.5, 0.0}
\definecolor{posmedium}{rgb}{0.0, 0.7, 0.0}
\definecolor{posweak}{rgb}{0.5, 0.8, 0.5}
\definecolor{negweak}{rgb}{0.8, 0.5, 0.5}
\definecolor{negmedium}{rgb}{0.7, 0.0, 0.0}
\definecolor{negstrong}{rgb}{0.5, 0.0, 0.0}

\newcommand{\colorcoef}[1]{%
  \ifdim #1 pt > 0.2pt
    \textcolor{posstrong}{#1}%
  \else\ifdim #1 pt > 0.1pt
    \textcolor{posmedium}{#1}%
  \else\ifdim #1 pt > 0.0pt
    \textcolor{posweak}{#1}%
  \else\ifdim #1 pt > -0.1pt
    \textcolor{negweak}{#1}%
  \else\ifdim #1 pt > -0.2pt
    \textcolor{negmedium}{#1}%
  \else
    \textcolor{negstrong}{#1}%
  \fi\fi\fi\fi\fi
}

\newcommand{\pctchange}[1]{\small\textcolor{gray}{(#1\%)}\normalsize}

\newcommand{\contribution}[2]{%
  #1 $\times$ \colorcoef{#2}%
}

\begin{document}
\maketitle

\begin{abstract}
We present a \emph{training-free} method to transplant tokenizers in pretrained large language models (LLMs) by reconstructing unseen token embeddings via \textbf{Orthogonal Matching Pursuit} (OMP). Specifically, we approximate each out-of-vocabulary token as a sparse linear combination of shared tokens, in two phases: first, compute each new token's representation in the donor embedding space with a small dictionary of shared anchor tokens, then transfer these same sparse coefficients back into the base model's embedding space.

On two challenging cross-tokenizer tasks---Llama$\to$Mistral NeMo (12B) and Qwen$\to$Llama (1B)---we show that OMP achieves best zero-shot preservation of the base model's performance across multiple benchmarks, while other zero-shot approaches degrade significantly. Compared to baselines (zero-init, mean-init, and existing approaches like WECHSEL, FOCUS, ZETT), OMP consistently achieves the best overall performance, effectively bridging large tokenizer discrepancies without gradient updates. Our analysis further identifies mismatched numerical tokenization schemes as a critical challenge for preserving mathematical reasoning capabilities. This technique enables direct reuse of pretrained model weights with new tokenizers, facilitating cross-tokenizer knowledge distillation, speculative decoding, ensembling, merging, and domain-specific vocabulary adaptations. We integrate our method into the open-source \texttt{mergekit-tokensurgeon} tool for \emph{post hoc} vocabulary realignment.
\end{abstract}

\section{Introduction}
Large language models (LLMs) are typically constrained by the tokenizer chosen during pretraining, which defines a fixed vocabulary for processing text. However, these vocabularies are \emph{not} universal; they may sub-optimally tokenize other languages, dialects, or domains~\citep{rust2021good,dobler-de-melo-2023-focus}. While replacing a model's tokenizer post-training is highly desirable in many scenarios, doing so without performance degradation remains a formidable technical challenge.

The core difficulty lies in embedding initialization for new tokens. Naive initialization can significantly degrade performance~\citep{minixhofer-etal-2022-wechsel,gee-etal-2022-fast}, while continuing pretraining to learn new embeddings can be prohibitively expensive.~\citep{gpt3,artetxe-etal-2020-cross} Meanwhile, existing zero-shot heuristics yield uneven results and can cause large performance drops on tasks like question answering or reasoning~\citep{dobler-de-melo-2023-focus,hewitt2021initializing}, especially when fundamental representational structures, like numerical tokenization schemes, differ between models.

This vocabulary mismatch creates practical barriers in several important scenarios. In \textbf{speculative decoding}~\citep{leviathan2023fast,ankner2024hydra}, a small model proposes partial outputs for a larger model to verify, requiring aligned tokenizers for direct logit comparison. \textbf{Knowledge distillation}~\citep{hinton2015distilling} similarly requires matched vocabularies when transferring token-level supervision from teacher to student~\citep{boizard2024towards}. \textbf{Model ensembling} or \textbf{merging}~\citep{goddard-etal-2024-arcees} presupposes that tokens map to the same embedding dimensions, with different tokenizers blocking direct output combination. An effective method for \emph{post hoc} tokenizer transfer---without retraining---would unlock these capabilities, greatly expanding the uses for existing models.

In this paper, we propose a fully \emph{training-free} approach to ``tokenizer transplantation,'' where we reconstruct embeddings for the \emph{new} vocabulary in the base model's embedding space using \textbf{Orthogonal Matching Pursuit} (OMP). Our key insight is to represent each new token's \emph{donor} embedding as a sparse combination of \emph{shared} token embeddings. Transplantation then becomes straightforward: we replicate those same sparse coefficients in the base model's embedding space. This yields a new vocabulary embedding matrix aligned with the \emph{new} tokenizer but consistent with the \emph{base model}'s embedding geometry.

We evaluate OMP-based transplantation in two cross-tokenizer experiments: Llama$\to$Mistral NeMo (12B) and Qwen$\to$Llama (1B). In both scenarios, OMP preserves model performance across classification, reasoning, and perplexity benchmarks far better than simple heuristics (zero, mean) or other zero-shot token-initialization approaches. We integrate this strategy into the \texttt{mergekit-tokensurgeon} open-source toolkit, enabling immediate usage in cross-tokenizer knowledge distillation, speculative decoding pipelines, or domain-vocabulary expansions, with no retraining required.

\paragraph{Contributions.}
\begin{enumerate}[noitemsep,leftmargin=*]
    \item We propose a \emph{training-free} tokenizer transplant method using Orthogonal Matching Pursuit in embedding space.
    \item We demonstrate superior zero-shot performance over existing approaches, evaluated on Llama$\to$Mistral NeMo and Qwen$\to$Llama tasks.
    \item We identify and analyze the significant negative impact of mismatched numerical tokenization schemes on mathematical reasoning performance, highlighting a key limitation of zero-shot transplantation.
    \item We highlight how OMP-based transplantation resolves key bottlenecks in cross-tokenizer use cases (e.g., teacher-student distillation, speculative decoding, or domain expansions).
    \item We provide an efficient incremental-QR OMP implementation and release it in \texttt{mergekit-tokensurgeon} for reproducibility.
\end{enumerate}

\section{Related Work}

\paragraph{Embedding Initialization for New Vocabularies.}
Assigning embeddings to new or unseen tokens is a well-explored problem in model adaptation. Naive (random/zero) initialization typically degrades performance substantially~\citep{hewitt2021initializing}. More sophisticated approaches often compute embeddings by \emph{averaging} or \emph{combining} subword pieces from the old vocabulary~\citep{minixhofer-etal-2022-wechsel,gee-etal-2022-fast}. For example, \citet{minixhofer-etal-2022-wechsel} propose WECHSEL, which uses bilingual word embeddings to map subword pieces from one language to another, then continues pretraining to refine the new embeddings. Similarly, \citet{dobler-de-melo-2023-focus} (FOCUS) and \citet{Ostendorff2023clp} (CLPTransfer) base their initialization on overlapping tokens, weighting them to approximate the new token's distribution. These methods yield better initial embeddings but often still rely on some continued training to recover the original model performance. In contrast, our method requires no further training for a high quality result. Additionally, OMP produces signed coefficients and therefore explores a larger linear subspace instead of restricting to the convex hull of anchors.

\paragraph{Zero-Shot Tokenizer Transfer.}
ZeTT~\citep{minixhofer2024zeroshottokenizertransfer} trains a hypernetwork for arbitrary tokenizers, applicable zero-shot after meta-training. Other tools~\citep{jukofyork2025transplant, goddard-etal-2024-arcees} copy or average existing embeddings. Our work aligns with these methods' goals of minimal or no additional training, but uses \emph{Orthogonal Matching Pursuit} (OMP) to anchor new token embeddings in a sparse set of existing embeddings. This approach yields more accurate approximations, especially with large vocabulary differences, and \emph{does not require a meta-trained hypernetwork}.

\paragraph{Sparse Approximation Techniques.}
Orthogonal Matching Pursuit (OMP)~\citep{pati1993orthogonal} is a standard technique for sparse reconstruction, used widely in signal processing and dictionary learning. While it has seen limited use in NLP\citep{menon2016clustering,skianis2018orthogonal}, it has not to our knowledge been adopted for cross-tokenizer embedding transfer. Our approach is the first to apply OMP as a solution to cross-tokenizer embedding alignment. 

\section{Methodology}
\label{sec:method}

\subsection{Problem Setup and Notation}
Let $\mathcal{M}_\mathrm{base}$ be a pretrained base model with vocabulary $V_\mathrm{base}$ and embedding matrix $E^\mathrm{(base)}\in\mathbb{R}^{|V_\mathrm{base}|\times d_\mathrm{base}}$. We want to \emph{replace} its vocabulary with $V_\mathrm{donor}$ from another model $\mathcal{M}_\mathrm{donor}$, which has an embedding matrix $E^\mathrm{(donor)}\in\mathbb{R}^{|V_\mathrm{donor}|\times d_\mathrm{donor}}$. Note that $d_\mathrm{donor}$ may differ from $d_\mathrm{base}$. We denote $V_\cap = V_\mathrm{base}\cap V_\mathrm{donor}$ as the set of shared (overlapping) tokens.

\paragraph{Case 1: Shared Tokens.}
If $t\in V_\cap$, we simply copy $e_t^\mathrm{(base)}$ to the new embedding matrix $E^\mathrm{(new)}$.

\paragraph{Case 2: Unseen Tokens.}
When $t\notin V_\mathrm{base}$, we have a \emph{donor embedding} $e_t^\mathrm{(donor)}$ but no corresponding $e_t^\mathrm{(base)}$. We must approximate $e_t^\mathrm{(base)}$ by referencing anchor tokens that exist in both vocabularies. In other words:
\[
e_t^\mathrm{(donor)} \approx \sum_{j \in A} \alpha_j\, e_j^\mathrm{(donor)},\quad A\subseteq V_\cap,\quad |A|\le k,
\]
and we then place
\[
e_t^\mathrm{(new)} \;=\; \sum_{j\in A} \alpha_j\, e_j^\mathrm{(base)}.
\]
Here, $k$ is a chosen sparsity level (e.g., 8, 32, 64), and $A$ is the small set of anchor tokens chosen by an OMP solver in the donor's embedding space. By applying those coefficients in the base model's space, we create the new embedding $e_t^\mathrm{(new)}$. As only the unit-agnostic coefficient vector $\alpha$ is transferred, there is no dependence on matching embedding dimensionality. 

\subsection{Orthogonal Matching Pursuit}

Orthogonal Matching Pursuit~\citep{pati1993orthogonal} is a greedy algorithm that builds a sparse representation of a target vector $v$ with respect to columns in a dictionary $\Phi$. In our setting:
\[
v = e_t^\mathrm{(donor)} \in \mathbb{R}^{d_\mathrm{donor}}, 
\quad
\Phi = \bigl[e_j^\mathrm{(donor)}\bigr]_{j\in V_\cap} \in \mathbb{R}^{d_\mathrm{donor}\times |V_\cap|}.
\]
We want 
\[
\min_{x\in \mathbb{R}^{|V_\cap|}} \|v - \Phi x\|_2 
\quad\text{subject to}\quad \|x\|_0\le k.
\]
OMP iteratively selects the dictionary column most correlated with the current residual, updates the partial least-squares solution, and refines the residual. After $k$ steps, we have a small set of anchor indices \(\Lambda\) plus nonzero coefficients \(x_\Lambda\). Algorithm~\ref{alg:omp} shows the procedure.

\begin{algorithm}[t!]
\caption{OMP for donor-space approximation}
\label{alg:omp}
\begin{algorithmic}[1]
\REQUIRE Dictionary $\Phi=\bigl[e_j^{(\mathrm{donor})}\bigr]_{j\in V_\cap}\in \mathbb{R}^{d\times |V_\cap|}$, target $v=e_t^{(\mathrm{donor})}\in \mathbb{R}^d$, sparsity $k$.
\ENSURE  A set of $k$ indices $\Lambda$ and sparse coeffs $x\in \mathbb{R}^{|V_\cap|}$.
\STATE Initialize $\Lambda \leftarrow \varnothing$, $r \leftarrow v$.
\FOR{$i=1 \ldots k$}
    \STATE $\displaystyle \lambda^* \leftarrow \arg\max_{j\in V_\cap} \Big|\langle r,\phi_j\rangle\Big|$
    \STATE $\Lambda \leftarrow \Lambda \cup\{\lambda^*\}$
    \STATE Solve $x_\Lambda = \arg\min_{x_\Lambda}\|\;v - \Phi_\Lambda x_\Lambda\|_2^2$ 
    \STATE $r \leftarrow v - \Phi_\Lambda x_\Lambda$
\ENDFOR
\STATE \textbf{return} $(\Lambda, x)$ where $x$ is zero outside $\Lambda$
\end{algorithmic}
\end{algorithm}

\paragraph{Efficient Implementation with Incremental QR Decomposition.}
Solving a least-squares problem each iteration can be expensive. We adopt an incremental QR factorization of $\Phi_\Lambda$ to update solutions more efficiently.

Rather than solving the least-squares problem from scratch at each iteration, we maintain a QR decomposition of the chosen columns of $\Phi$ and update it incrementally as we add new columns. Given $\Phi_{t-1} = Q_{t-1}R_{t-1}$, when we add column $\phi_{\lambda_t}$, we: (1)~compute $\hat{q} = \phi_{\lambda_t} - Q_{t-1}Q_{t-1}^T\phi_{\lambda_t}$ (the component orthogonal to existing columns); (2)~normalize to get $q_t = \hat{q}/\|\hat{q}\|_2$; and (3)~extend $Q_{t-1}$ with $q_t$ and update $R_{t-1}$ accordingly.
For numerical stability we re-orthogonalize at fixed intervals, but note that the atom selection of OMP is itself well-suited to avoiding catastrophic cancellation---results with and without re-orthogonalization differ only up to floating point precision.

This approach reduces the per-iteration complexity from $O(td^2)$ to $O(td)$, making the algorithm practical for large numbers of anchor points.

\paragraph{Forming the Base Embedding.}
After obtaining the sparse coefficients $x_\Lambda$, we place
\[
e_t^\mathrm{(new)} \;=\; \sum_{j\in\Lambda} x_j\, e_j^\mathrm{(base)}.
\]
We repeat this for all unseen tokens $t\notin V_\mathrm{base}$, while shared tokens $t\in V_\cap$ simply copy their existing base embeddings. Finally, we replace the original embedding matrix $E^\mathrm{(base)}$ with $E^\mathrm{(new)}$, ensuring the model now matches the donor tokenizer's vocabulary \emph{and} ID mapping.

\subsection{Geometric Justification via Approximate Orthogonal Alignment}
\label{sec:theory_justification}

Our method's rationale relies on two principles from embedding alignment and sparse recovery:

\paragraph{Approximate Orthogonal Equivalence.}
Independently trained embedding spaces (e.g., for words or LLM tokens~\citep{kulshreshtha-etal-2020-cross,lee2025sharedgloballocalgeometry}) have been shown to often align via an approximately orthogonal transformation \(U \in \mathbb{R}^{d_\mathrm{base}\times d_\mathrm{donor}}\) (where \(UU^\top\approx I_{d_\mathrm{base}}\), \(U^\top U\approx I_{d_\mathrm{donor}}\)) on their shared subspace~\citep{conneau2018word}. Thus, for \(j\in V_\cap\), \(U\,e_j^{(\mathrm{donor})} \approx e_j^{(\mathrm{base})}\), meaning \(U\) applied to linear combinations of donor embeddings approximates the same combinations in base space.

\paragraph{Sparse Reconstruction Stability.}
Orthogonal Matching Pursuit (OMP)~\citep{pati1993orthogonal,tropp2007signal} reliably reconstructs a target vector from a few dictionary atoms if the dictionary exhibits mild incoherence or Restricted Isometry Properties. If \(e_t^{(\mathrm{donor})} \approx \sum_{j\in A}\alpha_j\,e_j^{(\mathrm{donor})}\), OMP finds \(\{\alpha_j\}\) with provably small error \(\lVert e_t^{(\mathrm{donor})} - \sum_{j\in A}\alpha_j\,e_j^{(\mathrm{donor})} \rVert_2\).

\paragraph{Putting It Together.}
Combining these, if \(e_t^{(\mathrm{donor})} \approx \sum_{j\in A}\alpha_j\,e_j^{(\mathrm{donor})}\), applying \(U\) yields:
\[
U\,e_t^{(\mathrm{donor})}
\;\approx\;
\sum_{j\in A}\alpha_j\,U\,e_j^{(\mathrm{donor})}
\;\approx\;
\sum_{j\in A}\alpha_j\,e_j^{(\mathrm{base})}
=
e_t^{(\mathrm{new})}.
\]
Transplanting OMP-derived coefficients into the base model's embedding space thus implicitly performs a least-squares projection and an approximate orthogonal alignment.

This provides a conceptual justification for our method's effectiveness, and though we do not prove (or seek to prove) this alignment universally exists, empirical results suggest it holds in practice.

\section{Experiments and Results}
\label{sec:experiments}

We evaluate our OMP-based approach on two cross-tokenizer settings:
\begin{itemize}[leftmargin=*]
    \item \textbf{Llama$\to$Mistral NeMo (12B):} Llama 3's \citep{grattafiori2024llama} $\sim$128k vocabulary transplanted into a 12B Mistral NeMo, which has $\sim$131k tokens. Overlap: $\sim$71k tokens ($\sim54\%$).
    \item \textbf{Qwen$\to$Llama (1B):} Qwen 2.5's \citep{yang2024qwen2} $\sim$152k vocabulary transplanted into a 1B-parameter Llama 3 model with a $\sim$128k vocabulary. Overlap: $\sim$110k tokens ($\sim86\%$).
\end{itemize}

In each case, we compare:
\begin{enumerate}[noitemsep,leftmargin=*]
    \item \textit{OMP} (our method) at various $k$ in \{8, 16, 32, 64, 256\}.
    \item \textit{ZeroEmbed}: all unseen tokens get zero vectors.
    \item \textit{MeanEmbed}: all unseen tokens get the mean embedding of the base vocabulary.
    \item \textit{Published approaches} (ZETT, FOCUS, WECHSEL, CLPTransfer) applied strictly zero-shot (i.e., no additional training) using the official released implementations.
\end{enumerate}

We evaluate perplexity on \texttt{WikiText}, plus classification or QA accuracy on \texttt{MMLU}~\citep{hendrycks2020measuring}, \texttt{ARC}~\citep{clark2018think}, \texttt{GSM8K}~\citep{cobbe2021training}, \texttt{LAMBADA}~\citep{paperno2016lambada}, \texttt{XNLI}~\citep{conneau2018xnli}, and \texttt{Paws-X}~\citep{yang2019paws}, among others, using Eleuther AI's LM Evaluation Harness~\citep{eval-harness}. For published baselines (such as ZETT, FOCUS, WECHSEL, and CLPTransfer), we applied only their proposed embedding initialization techniques, omitting any subsequent fine-tuning or continued pre-training steps suggested in their original methodologies, to ensure a direct comparison of zero-shot initialization quality. We additionally compare post-training results given identical token budgets in Section~\ref{sec:cpt_results}.

Table \ref{tab:stderr_summary} summarizes the evaluation settings and typical standard errors (stderr) associated with the primary metric for key benchmarks, as computed by the evaluation harness via bootstrapping \citep{eval-harness}. These errors quantify uncertainty stemming from the finite size of the evaluation dataset sample. Since the standard errors were consistently small relative to the performance differences observed between methods and did not affect the relative rankings or main conclusions, we omit them from the main results tables (Tables \ref{tab:llama-mistral-results}, \ref{tab:llama_qwen_benchmarks}, \ref{tab:nemo_qwen_numeric}, \ref{table:cpt-evals}) for clarity.

\begin{table}[ht]
\centering
\caption{Evaluation settings and representative standard errors (stderr) for key benchmarks.}
\label{tab:stderr_summary}
\begin{tabular}{lccc}
\toprule
\textbf{Benchmark} & \textbf{Metric} & \textbf{Few-shot} & \textbf{StdErr} \\
\midrule
MMLU & acc (agg.) & 0 & 0.004 \\
ARC Challenge & acc\_norm & 0 & 0.014 \\
GSM8K & exact\_match (flex) & 5 & 0.013 \\
XNLI & acc (agg.) & 0 & 0.003 \\
AGIEval & acc (agg.) & 0 & 0.005 \\
Lambada (EN) & acc & 0 & 0.006 \\
WikiText & bits\_per\_byte & 0 & N/A\textsuperscript{*} \\
\bottomrule
\multicolumn{4}{l}{\footnotesize\textsuperscript{*} LM Eval Harness does not report bootstrapped stderr for perplexity metrics.} \\
\multicolumn{4}{l}{\footnotesize `(flex)' refers to flexible answer extraction; `(agg.)' is aggregated score.}
\end{tabular}
\end{table}

\subsection{Llama$\to$Mistral NeMo (12B)}

In Llama$\to$Mistral NeMo, the baseline NeMo model has strong MMLU accuracy (64.5\%). Table~\ref{tab:llama-mistral-results} shows that OMP with $k=64$ preserves MMLU at 62.2\% (-3.6\%), while naive baselines drop to $\approx 58\%$ (-9.3\%). While all zero-shot transplants significantly degrade mathematical tasks, OMP remains the most consistent overall.

\begin{table}[ht]
\centering
\caption{Llama$\to$Mistral NeMo results. Relative performance changes shown in parentheses.}
\label{tab:llama-mistral-results}
\begin{tabular}{lccccccc}
\toprule
\textbf{Model} & \textbf{MMLU} & \textbf{ARC-C} & \textbf{XNLI} & \textbf{GSM8K} & \textbf{AGIEval} & \textbf{Lambada} & \textbf{WikiText} \\
&  &  &  &  &  & \textbf{(EN)} & \textbf{(Bits/Byte) $\downarrow$} \\
\midrule
\textbf{Baseline (Mistral NeMo)} & 0.6452 & 0.5811 & 0.4367 & 0.5588 & 0.3752 & 0.7819 & 0.5296 \\
\midrule
\textbf{OMP-K8} & 0.6129 & 0.5179 & 0.3782 & 0.1289 & 0.3137 & 0.6897 & 0.7798 \\
    & \pctchange{-5.00} & \pctchange{-10.87} & \pctchange{-13.39} & \pctchange{-76.93} & \pctchange{-16.38} & \pctchange{-11.79} & \pctchange{+47.24} \\
\textbf{OMP-K32} & 0.6158 & 0.5239 & 0.3780 & 0.1296 & 0.3146 & 0.6905 & 0.8048 \\
    & \pctchange{-4.56} & \pctchange{-9.84} & \pctchange{-13.44} & \pctchange{-76.80} & \pctchange{-16.15} & \pctchange{-11.69} & \pctchange{+51.95} \\
\textbf{OMP-K64} & 0.6222 & 0.5273 & 0.3780 & 0.1463 & 0.3158 & 0.6895 & 0.7417 \\
    & \pctchange{-3.57} & \pctchange{-9.25} & \pctchange{-13.43} & \pctchange{-73.81} & \pctchange{-15.83} & \pctchange{-11.81} & \pctchange{+40.05} \\
\midrule
\textbf{FOCUS} & 0.2299 & 0.2585 & 0.3286 & 0.0076 & 0.2447 & 0.0000 & 5.8441 \\
    & \pctchange{-64.37} & \pctchange{-55.51} & \pctchange{-24.74} & \pctchange{-98.64} & \pctchange{-34.78} & \pctchange{-100.00} & \pctchange{+1003.47} \\
\textbf{ZeroEmbed} & 0.5852 & 0.4949 & 0.3689 & 0.0334 & 0.2905 & 0.6598 & 0.9881 \\
    & \pctchange{-9.29} & \pctchange{-14.83} & \pctchange{-15.53} & \pctchange{-94.03} & \pctchange{-22.57} & \pctchange{-15.61} & \pctchange{+86.58} \\
\textbf{MeanEmbed} & 0.5936 & 0.4966 & 0.3693 & 0.0523 & 0.2952 & 0.6670 & 0.8993 \\
    & \pctchange{-8.00} & \pctchange{-14.54} & \pctchange{-15.43} & \pctchange{-90.64} & \pctchange{-21.31} & \pctchange{-14.69} & \pctchange{+69.81} \\
\bottomrule
\end{tabular}
\end{table}

\subsection{Qwen$\to$Llama (1B) Results}

The Qwen$\to$Llama transplantation presents an interesting scenario due to its very high token overlap (~86\%), which includes almost all English tokens relevant to benchmarks like MMLU and ARC. Table~\ref{tab:llama_qwen_benchmarks} shows key metrics. Here the resilience of \textit{ZeroEmbed} and \textit{MeanEmbed} suggests a key factor is avoiding interference with these shared embeddings by new token initializations. OMP navigates this high-overlap scenario adeptly. It preserves near-baseline performance for MMLU (36.40\% vs. 36.73\% baseline) and ARC (36.26\% vs. 36.26\% baseline), and performs strongly on Belebele, with only a slight increase in perplexity on WikiText (0.7159 vs. 0.6605 bits/byte). This performance outperforms all simpler heuristics. For GSM8K, again all zero-shot methods degrade severely; OMP is still competitive.

\begin{table}[ht]
\centering
\caption{Qwen$\to$Llama (1B) tokenizer-transplant performance.}
\label{tab:llama_qwen_benchmarks}
\begin{tabular}{lccccc}
\toprule
\textbf{Model} & \textbf{MMLU} & \textbf{ARC-C} & \textbf{GSM8K} & \textbf{Belebele} & \textbf{WikiText} \\
&  &  &  &  & \textbf{(Bits/Byte) $\downarrow$} \\
\midrule
\textbf{Baseline (Llama)} & 0.3673 & 0.3626 & 0.0675 & 0.2813 & 0.6605 \\
\midrule
\textbf{OMP-K8} & 0.3641 & 0.3626 & 0.0144 & 0.2687 & 0.7160 \\
    & \pctchange{-0.85} & \pctchange{+0.00} & \pctchange{-78.65} & \pctchange{-4.46} & \pctchange{+8.40} \\
\textbf{OMP-K32} & 0.3640 & 0.3626 & 0.0144 & 0.2693 & 0.7160 \\
    & \pctchange{-0.89} & \pctchange{+0.00} & \pctchange{-78.65} & \pctchange{-4.25} & \pctchange{+8.39} \\
\textbf{OMP-K64} & 0.3640 & 0.3626 & 0.0144 & 0.2704 & 0.7159 \\
    & \pctchange{-0.89} & \pctchange{+0.00} & \pctchange{-78.65} & \pctchange{-3.86} & \pctchange{+8.38} \\
\midrule
\textbf{CLPTransfer} & 0.2295 & 0.2858 & 0.0000 & 0.2299 & 6791.6044 \\
    & \pctchange{-37.52} & \pctchange{-21.18} & \pctchange{-100.00} & \pctchange{-18.26} & \pctchange{+$1\mathrm{e}{6}$} \\
\textbf{FOCUS} & 0.2695 & 0.3549 & 0.0174 & 0.2478 & 0.8870 \\
    & \pctchange{-26.62} & \pctchange{-2.12} & \pctchange{-74.16} & \pctchange{-11.91} & \pctchange{+34.29} \\
\textbf{WECHSEL} & 0.2314 & 0.2389 & 0.0099 & 0.2470 & 3.1852 \\
    & \pctchange{-36.98} & \pctchange{-34.12} & \pctchange{-85.39} & \pctchange{-12.17} & \pctchange{+382.24} \\
\textbf{ZETT} & 0.2634 & 0.3055 & 0.0000 & 0.2480 & 1.1426 \\
    & \pctchange{-28.27} & \pctchange{-15.76} & \pctchange{-100.00} & \pctchange{-11.82} & \pctchange{+72.99} \\
\textbf{ZeroEmbed} & 0.3640 & 0.3626 & 0.0144 & 0.2648 & 0.7169 \\
    & \pctchange{-0.89} & \pctchange{+0.00} & \pctchange{-78.65} & \pctchange{-5.87} & \pctchange{+8.53} \\
\textbf{MeanEmbed} & 0.3641 & 0.3626 & 0.0144 & 0.2577 & 0.7180 \\
    & \pctchange{-0.85} & \pctchange{+0.00} & \pctchange{-78.65} & \pctchange{-8.38} & \pctchange{+8.70} \\
\bottomrule
\end{tabular}
\end{table}

\subsection{Effect of $k$.}
We examined how the sparsity parameter $k$ affects performance across benchmarks. Our experiments tested values ranging from $k=8$ to $k=256$.

In the Qwen$\to$Llama experiment, we found minimal gains beyond $k=16$, with performance plateauing for most tasks around $k=32$--$64$. For instance, MMLU scores were 36.41\%, 36.40\%, and 36.41\% for $k=8$, $k=32$, and $k=256$, respectively.

In the Llama$\to$Mistral NeMo case, we observed more substantial improvements with higher $k$ values, particularly in mathematical tasks like GSM8K (12.89\% at $k=8$ vs. 14.63\% at $k=64$). However, increasing beyond $k=64$ yielded diminishing returns.

We recommend $k=64$ as offering the best balance of performance and efficiency.

\subsection{Numerical Performance Degradation.}

A striking observation across both the Llama$\to$Mistral NeMo (Table~\ref{tab:llama-mistral-results}) and Qwen$\to$Llama (Table~\ref{tab:llama_qwen_benchmarks}) experiments is the dramatic degradation in mathematical reasoning performance, particularly on GSM8K (drops of -73.8\% and -78.7\% for OMP-K64, respectively, compared to baselines). We hypothesize this stems from fundamental differences in numerical tokenization schemes. Mistral NeMo and Qwen employ single-digit tokenization (e.g., "1234" $\to$ "1", "2", "3", "4"), resulting in only 10 base numeric tokens. In contrast, Llama 3 uses triplet-based chunking (e.g., "1234" $\to$ "123", "4"), leading to a vastly larger numerical vocabulary (1110 dedicated number tokens).

During transplantation between models with mismatched schemes (like Qwen$\to$Llama or Llama$\to$Mistral NeMo), the vast majority of one model's numeric tokens lack direct equivalents and must be approximated via OMP using potentially non-numeric anchors. This likely introduces systematic distortions in the model's learned numerical representations and operations, which may rely on specific geometric structures tied to the pretraining tokenization scheme. For example, if models represent numbers on a 'generalized helix'~\citep{kantamneni2025language} structure sensitive to the sequence of numeric tokens, then approximating embeddings across mismatched schemes (like reconstructing Llama's triplet '123' using Qwen's '1', '2', '3' anchors via OMP) could fail to preserve this geometric arrangement and thus impair arithmetic capabilities. Tellingly, transplanted models in these mismatched scenarios frequently produce single-digit answers to multi-digit problems, suggesting that mathematical reasoning capabilities are tightly coupled to the specific tokenization patterns encountered during pretraining.

To validate this hypothesis, we conducted an additional experiment transplanting the Mistral NeMo tokenizer into the Qwen 2.5 7B model. Crucially, both models utilize the \emph{same} single-digit numerical tokenization scheme, although their overall vocabularies differ. As shown in Table~\ref{tab:nemo_qwen_numeric} the GSM8K performance drop for OMP-K64 in this Mistral NeMo$\to$Qwen transplantation was only \textbf{-5.6\%} (from 82.6\% to 78.0\%). This contrasts sharply with the >70\% degradation observed when numeric schemes were mismatched.

\begin{table}[ht]
\centering
\caption{Benchmark results for Mistral NeMo$\to$Qwen}
\label{tab:nemo_qwen_numeric}
\begin{tabular}{lccccccc}
\toprule
\textbf{Model} & \textbf{MMLU} & \textbf{ARC-C} & \textbf{XNLI} & \textbf{GSM8K} & \textbf{AGIEval} & \textbf{Lambada} & \textbf{WikiText} \\
&  &  &  &  &  & \textbf{(EN)} & \textbf{(Bits/Byte) $\downarrow$} \\
\midrule
\textbf{Baseline (Qwen)} & 0.7196 & 0.5137 & 0.4344 & 0.8264 & 0.5639 & 0.7173 & 0.5847 \\
\midrule
\textbf{OMP-K64} & 0.7064 & 0.4872 & 0.3769 & 0.7801 & 0.4423 & 0.6396 & 0.6781 \\
 & \pctchange{-1.83} & \pctchange{-5.15} & \pctchange{-13.22} & \pctchange{-5.60} & \pctchange{-21.56} & \pctchange{-10.82} & \pctchange{+15.97} \\
\midrule
\textbf{MeanEmbed} & 0.6839 & 0.4693 & 0.3560 & 0.7900 & 0.4288 & 0.6309 & 0.7524 \\
 & \pctchange{-4.95} & \pctchange{-8.64} & \pctchange{-18.04} & \pctchange{-4.40} & \pctchange{-23.96} & \pctchange{-12.04} & \pctchange{+28.69} \\
\textbf{ZeroEmbed} & 0.6874 & 0.4804 & 0.3542 & 0.7779 & 0.4278 & 0.6299 & 0.7722 \\
 & \pctchange{-4.46} & \pctchange{-6.48} & \pctchange{-18.47} & \pctchange{-5.87} & \pctchange{-24.13} & \pctchange{-12.18} & \pctchange{+32.07} \\
\midrule
\textbf{CLPTransfer} & 0.2297 & 0.2474 & 0.3326 & 0.0038 & 0.2456 & 0.0000 & 5.8987 \\
 & \pctchange{-68.08} & \pctchange{-51.83} & \pctchange{-23.44} & \pctchange{-99.54} & \pctchange{-56.44} & \pctchange{-100.00} & \pctchange{+908.82} \\
\bottomrule
\end{tabular}
\end{table}

This result strongly supports our hypothesis: when numeric tokenization schemes are aligned, OMP successfully preserves mathematical performance to a much greater degree. While OMP effectively bridges semantic representations for general text tokens, the structural differences in representing numbers pose a unique challenge that significantly impacts arithmetic and quantitative reasoning unless the underlying numeric tokenization strategy is preserved or highly similar. Specialized handling of numeric tokens during transplantation may be required to fully retain mathematical abilities across arbitrary tokenizer pairs.
\subsection{Continued Pre-Training}
\label{sec:cpt_results}

Although our approach does not require additional training, we note that partial fine-tuning or domain adaptation can boost performance further. Our experiments confirm that even a small amount of continued training helps recover performance on sensitive tasks (like GSM8K), but zero-shot OMP alone already outperforms other zero-shot heuristics.

The models were fine-tuned for a single epoch on a two-billion token subset of the DCLM~\citep{li2024datacomplm} dataset. All baseline models were tuned with the AdamW optimizer and a learning rate of 5e-6. OMP required a much lower learning rate of 4e-7. This might suggest that the OMP-initialized embeddings are already well-positioned and sensitive to larger updates. The results of this experiment are shown in Table~\ref{table:cpt-evals}. We note the unexpected slight degradation of OMP on XNLI post-CPT. While the exact cause is unclear, given XNLI's cross-lingual nature and the English-only CPT dataset, this might suggest that CPT subtly disrupted the initial zero-shot alignment for this specific task, potentially due to the sensitivity of the OMP-initialized embeddings (as evidenced by the required lower learning rate).

\begin{table}[ht]
    \centering
    \caption{Performance comparison of tokenizer transplantation methods for the Qwen$\to$Llama pair with and without continued pretraining on the DCLM dataset. Percentages in gray show relative performance change compared to the baseline model. Lower values are better for WikiText (Bits/Byte); higher values are better for all other metrics.}
    \label{table:cpt-evals}
    \begin{tabular}{lcccccc}
    \toprule
    \textbf{Model} & \textbf{MMLU} & \textbf{XNLI} & \textbf{AGIEval} & \textbf{LAMBADA} & \textbf{Belebele} & \textbf{WikiText} \\
    & & & & \textbf{(EN)} & & \textbf{(Bits/Byte)} $\downarrow$ \\
    \midrule
    \textbf{Baseline} & 0.3673 & 0.4086 & 0.2690 & 0.6204 & 0.2813 & 0.6605 \\
    \midrule
    \multicolumn{7}{l}{\textit{Zero-Shot}} \\
    \midrule
    \textbf{ZETT} & 0.2634 & 0.3437 & 0.2487 & 0.3460 & 0.2480 & 1.1426 \\
    & \pctchange{-28.3} & \pctchange{-15.9} & \pctchange{-7.6} & \pctchange{-44.2} & \pctchange{-11.8} & \pctchange{+73.0} \\
    \textbf{FOCUS} & 0.2695 & 0.3578 & 0.2577 & 0.5936 & 0.2478 & 0.8870 \\
    & \pctchange{-26.6} & \pctchange{-12.4} & \pctchange{-4.2} & \pctchange{-4.3} & \pctchange{-11.9} & \pctchange{+34.3} \\
    \textbf{OMP-K64} & 0.3640 & 0.3430 & 0.2605 & 0.6200 & 0.2704 & 0.7159 \\
    & \pctchange{-0.9} & \pctchange{-16.1} & \pctchange{-3.1} & \pctchange{-0.1} & \pctchange{-3.9} & \pctchange{+8.4} \\
    \midrule
    \multicolumn{7}{l}{\textit{With Continued Pre-Training}} \\
    \midrule
    \textbf{ZETT-DCLM-2B} & 0.3151 & 0.3784 & 0.2638 & 0.6115 & 0.2725 & 0.6819 \\
    & \pctchange{-14.2} & \pctchange{-7.4} & \pctchange{-1.9} & \pctchange{-1.4} & \pctchange{-3.1} & \pctchange{+3.2} \\
    \textbf{FOCUS-DCLM-2B} & 0.3444 & 0.3750 & 0.2579 & 0.6198 & 0.2559 & 0.6733 \\
    & \pctchange{-6.2} & \pctchange{-8.2} & \pctchange{-4.1} & \pctchange{-0.1} & \pctchange{-9.0} & \pctchange{+1.9} \\
    \textbf{OMP-K64-DCLM-2B} & 0.3725 & 0.3388 & 0.2604 & 0.6185 & 0.2724 & 0.6730 \\
    & \pctchange{+1.4} & \pctchange{-17.1} & \pctchange{-3.2} & \pctchange{-0.3} & \pctchange{-3.2} & \pctchange{+1.9} \\
    \bottomrule
    \end{tabular}
\end{table}

\subsection{Computational Efficiency.}

Beyond performance metrics, the practical utility of tokenizer transplantation depends on its computational cost. We measured the approximate execution time for the Qwen$\to$Llama (1B) task, involving the approximation of $\sim$41,000 tokens. OMP proves highly efficient: on a single H100 GPU, it required only 38 seconds (with $k=8$) and 74 seconds (with $k=32$). Naive methods like ZeroEmbed and MeanEmbed had negligible computational cost (effectively instantaneous on a CPU). In contrast, other non-trivial zero-shot approaches were significantly more demanding: CLPTransfer took approximately 9 hours on a CPU, and FOCUS required a similar duration (exact time not recorded but observed to be comparable). ZeTT involves a substantial one-time meta-training cost (39 hours on an 8x H100 GPU node in our setup) followed by a fast per-model application step (2 minutes on a single H100 GPU for this task). This comparison highlights OMP's advantage as a truly \textit{post hoc}, lightweight solution for rapid deployment without extensive pre-computation or runtimes.

\section{Discussion}

\paragraph{Why Does OMP Work Well?}
Word-embedding spaces exhibit approximate local linearity and analogical structure~\citep{mikolov2013efficient} and large-language-model embeddings form tight semantic clusters in high dimensions~\citep{ethayarajh2019contextual}. OMP leverages both facts: by greedily selecting a \emph{signed}, $k$-sparse code of shared anchor embeddings, it captures the dominant semantic directions of an unseen token without any gradient updates.

\paragraph{Applications.}
\begin{enumerate}[noitemsep,leftmargin=*]
\item \textbf{Cross-tokenizer knowledge distillation.} Teacher and student often differ in vocabulary; by transplanting the teacher's tokenizer onto the student, we can directly apply cross-entropy or logit-based distillation with matched token IDs.
\item \textbf{Speculative decoding.}~\citep{leviathan2023fast} A smaller ``draft'' model must share a vocabulary with the larger ``verifier'' model. OMP-based transplantation allows arbitrary pairs of models to be used regardless of original vocabulary.
\item \textbf{Domain vocabulary expansions.} Instead of a full replacement, OMP can be used to initialize embeddings for new domain-specific tokens (e.g., medical, chemical, code, or multilingual) by reconstructing them from existing anchors. This adds minimal overhead and preserves existing performance.
\end{enumerate}

\paragraph{Limitations \& Future Work.}
Our approach depends on having at least some overlap in $V_\cap$; if none exists, the method cannot be applied---though as any modern byte-level or Unicode-complete tokenizer assigns \textit{some} ID to every code-point sequence, in practice $|V_\cap|>0$. Structurally different numeric tokenization schemes (such as digit clustering or right-to-left vs. left-to-right segmentation) may degrade math tasks significantly. Building on our finding of this critical limitation, a key direction for future work is the development of hybrid transplantation strategies. Such strategies would leverage OMP for general vocabulary while employing specialized handling for numeric tokens specifically designed to bridge these structural representational gaps. Other avenues include bridging strategies for near-disjoint vocabularies or applying other sparse coding techniques.

\subsection*{Broader Impacts and Ethical Considerations}
The primary motivation behind our work is to enhance the flexibility and reusability of pretrained language models for beneficial applications such as improved knowledge distillation, efficient speculative decoding, and domain-specific adaptations. By enabling training-free tokenizer transplantation, we aim to lower technical barriers and promote innovation in these areas.

However, as with any technology that increases the ease of model modification and interoperability, there are potential negative societal impacts to consider. While our method does not inherently create new malicious capabilities within a model, it could inadvertently lower the technical or computational barriers for adapting existing models for unintended or harmful purposes. For instance: 
\begin{itemize}[noitemsep,leftmargin=*]
    \item \textbf{Facilitating Adaptation of Problematic Models:} If a base model possesses capabilities that could be misused (e.g., generating sophisticated disinformation, exhibiting strong biases, or producing unsafe content), our technique could make it more efficient for malicious actors to adapt such a model to new vocabularies or integrate it into harmful pipelines. The training-free nature reduces the cost and expertise typically associated with such re-tokenization efforts.
    \item \textbf{Propagation of Biases and Harms:} The transplantation process directly transfers learned representations. If the base model contains unmitigated biases or safety flaws, these could be seamlessly propagated when its vocabulary is altered, potentially affecting new languages or domains targeted by the transplanted tokenizer if not carefully evaluated.
\end{itemize}
It must be emphasized that the ethical responsibilities associated with the deployment of LLMs remain with the developers and users. Our tool is a component that operates on existing models, and the onus is on the user to ensure that the base models are used responsibly and that any model resulting from tokenizer transplantation is thoroughly evaluated for safety, fairness, and its intended application before deployment. We advocate for continued research into robust evaluation techniques and safeguards for all language models, regardless of their tokenization scheme.

\section{Conclusion}

We introduce a \textbf{training-free} approach for tokenizer transplantation using \textit{Orthogonal Matching Pursuit} on shared-token embeddings. Our experiments show that OMP preserves perplexity and classification accuracy far better than naive heuristics, enabling convenient reuse of pretrained LLM weights under new tokenizers. This approach unlocks crucial applications (cross-tokenizer distillation, speculation, domain expansions) by eliminating vocabulary mismatches without any additional training. Our implementation is openly available in \texttt{mergekit-tokensurgeon}, inviting broader adoption and future enhancements. By highlighting the power of sparse approximation in embedding space, we hope to inspire further modular, post hoc enhancements for pretrained LLMs.

\bibliographystyle{plainnat}
\bibliography{references}

\begin{thebibliography}{35}
\providecommand{\natexlab}[1]{#1}
\providecommand{\url}[1]{\texttt{#1}}
\expandafter\ifx\csname urlstyle\endcsname\relax
  \providecommand{\doi}[1]{doi: #1}\else
  \providecommand{\doi}{doi: \begingroup \urlstyle{rm}\Url}\fi

\bibitem[Ankner et~al.(2024)Ankner, Parthasarathy, Nrusimha, Rinard, Ragan-Kelley, and Brandon]{ankner2024hydra}
Zachary Ankner, Rishab Parthasarathy, Aniruddha Nrusimha, Christopher Rinard, Jonathan Ragan-Kelley, and William Brandon.
\newblock Hydra: Sequentially-dependent draft heads for medusa decoding.
\newblock \emph{arXiv preprint arXiv:2402.05109}, 2024.

\bibitem[Artetxe et~al.(2020)Artetxe, Ruder, and Yogatama]{artetxe-etal-2020-cross}
Mikel Artetxe, Sebastian Ruder, and Dani Yogatama.
\newblock On the cross-lingual transferability of monolingual representations.
\newblock In Dan Jurafsky, Joyce Chai, Natalie Schluter, and Joel Tetreault, editors, \emph{Proceedings of the 58th Annual Meeting of the Association for Computational Linguistics}, pages 4623--4637, Online, July 2020. Association for Computational Linguistics.
\newblock \doi{10.18653/v1/2020.acl-main.421}.
\newblock URL \url{https://aclanthology.org/2020.acl-main.421/}.

\bibitem[Boizard et~al.(2024)Boizard, Haddad, Hudelot, and Colombo]{boizard2024towards}
Nicolas Boizard, Kevin~El Haddad, C{\'e}line Hudelot, and Pierre Colombo.
\newblock Towards cross-tokenizer distillation: the universal logit distillation loss for llms.
\newblock \emph{arXiv preprint arXiv:2402.12030}, 2024.

\bibitem[Brown et~al.(2020)Brown, Mann, Ryder, Subbiah, Kaplan, Dhariwal, Neelakantan, Shyam, Sastry, Askell, et~al.]{gpt3}
Tom Brown, Benjamin Mann, Nick Ryder, Melanie Subbiah, Jared~D Kaplan, Prafulla Dhariwal, Arvind Neelakantan, Pranav Shyam, Girish Sastry, Amanda Askell, et~al.
\newblock Language models are few-shot learners.
\newblock \emph{Advances in neural information processing systems}, 33:\penalty0 1877--1901, 2020.

\bibitem[Clark et~al.(2018)Clark, Cowhey, Etzioni, Khot, Sabharwal, Schoenick, and Tafjord]{clark2018think}
Peter Clark, Isaac Cowhey, Oren Etzioni, Tushar Khot, Ashish Sabharwal, Carissa Schoenick, and Oyvind Tafjord.
\newblock Think you have solved question answering? try arc, the ai2 reasoning challenge.
\newblock \emph{arXiv preprint arXiv:1803.05457}, 2018.

\bibitem[Cobbe et~al.(2021)Cobbe, Kosaraju, Bavarian, Chen, Jun, Kaiser, Plappert, Tworek, Hilton, Nakano, et~al.]{cobbe2021training}
Karl Cobbe, Vineet Kosaraju, Mohammad Bavarian, Mark Chen, Heewoo Jun, Lukasz Kaiser, Matthias Plappert, Jerry Tworek, Jacob Hilton, Reiichiro Nakano, et~al.
\newblock Training verifiers to solve math word problems.
\newblock \emph{arXiv preprint arXiv:2110.14168}, 2021.

\bibitem[Conneau et~al.(2018{\natexlab{a}})Conneau, Lample, Ranzato, Denoyer, and Jégou]{conneau2018word}
Alexis Conneau, Guillaume Lample, Marc'Aurelio Ranzato, Ludovic Denoyer, and Hervé Jégou.
\newblock Word translation without parallel data, 2018{\natexlab{a}}.
\newblock URL \url{https://arxiv.org/abs/1710.04087}.

\bibitem[Conneau et~al.(2018{\natexlab{b}})Conneau, Lample, Rinott, Williams, Bowman, Schwenk, and Stoyanov]{conneau2018xnli}
Alexis Conneau, Guillaume Lample, Ruty Rinott, Adina Williams, Samuel~R Bowman, Holger Schwenk, and Veselin Stoyanov.
\newblock Xnli: Evaluating cross-lingual sentence representations.
\newblock \emph{arXiv preprint arXiv:1809.05053}, 2018{\natexlab{b}}.

\bibitem[Dobler and de~Melo(2023)]{dobler-de-melo-2023-focus}
Konstantin Dobler and Gerard de~Melo.
\newblock {FOCUS}: Effective embedding initialization for monolingual specialization of multilingual models.
\newblock In Houda Bouamor, Juan Pino, and Kalika Bali, editors, \emph{Proceedings of the 2023 Conference on Empirical Methods in Natural Language Processing}, pages 13440--13454, Singapore, December 2023. Association for Computational Linguistics.
\newblock \doi{10.18653/v1/2023.emnlp-main.829}.
\newblock URL \url{https://aclanthology.org/2023.emnlp-main.829/}.

\bibitem[Ethayarajh(2019)]{ethayarajh2019contextual}
Kawin Ethayarajh.
\newblock How contextual are contextualized word representations? {C}omparing the geometry of {BERT}, {ELM}o, and {GPT}-2 embeddings.
\newblock In Kentaro Inui, Jing Jiang, Vincent Ng, and Xiaojun Wan, editors, \emph{Proceedings of the 2019 Conference on Empirical Methods in Natural Language Processing and the 9th International Joint Conference on Natural Language Processing (EMNLP-IJCNLP)}, pages 55--65, Hong Kong, China, November 2019. Association for Computational Linguistics.
\newblock \doi{10.18653/v1/D19-1006}.
\newblock URL \url{https://aclanthology.org/D19-1006/}.

\bibitem[Gao et~al.(2023)Gao, Tow, Abbasi, Biderman, Black, DiPofi, Foster, Golding, Hsu, Le~Noac'h, Li, McDonell, Muennighoff, Ociepa, Phang, Reynolds, Schoelkopf, Skowron, Sutawika, Tang, Thite, Wang, Wang, and Zou]{eval-harness}
Leo Gao, Jonathan Tow, Baber Abbasi, Stella Biderman, Sid Black, Anthony DiPofi, Charles Foster, Laurence Golding, Jeffrey Hsu, Alain Le~Noac'h, Haonan Li, Kyle McDonell, Niklas Muennighoff, Chris Ociepa, Jason Phang, Laria Reynolds, Hailey Schoelkopf, Aviya Skowron, Lintang Sutawika, Eric Tang, Anish Thite, Ben Wang, Kevin Wang, and Andy Zou.
\newblock A framework for few-shot language model evaluation, 12 2023.
\newblock URL \url{https://zenodo.org/records/10256836}.

\bibitem[Gee et~al.(2022)Gee, Zugarini, Rigutini, and Torroni]{gee-etal-2022-fast}
Leonidas Gee, Andrea Zugarini, Leonardo Rigutini, and Paolo Torroni.
\newblock Fast vocabulary transfer for language model compression.
\newblock In Yunyao Li and Angeliki Lazaridou, editors, \emph{Proceedings of the 2022 Conference on Empirical Methods in Natural Language Processing: Industry Track}, pages 409--416, Abu Dhabi, UAE, December 2022. Association for Computational Linguistics.
\newblock \doi{10.18653/v1/2022.emnlp-industry.41}.
\newblock URL \url{https://aclanthology.org/2022.emnlp-industry.41/}.

\bibitem[Goddard et~al.(2024)Goddard, Siriwardhana, Ehghaghi, Meyers, Karpukhin, Benedict, McQuade, and Solawetz]{goddard-etal-2024-arcees}
Charles Goddard, Shamane Siriwardhana, Malikeh Ehghaghi, Luke Meyers, Vladimir Karpukhin, Brian Benedict, Mark McQuade, and Jacob Solawetz.
\newblock Arcee`s {M}erge{K}it: A toolkit for merging large language models.
\newblock In Franck Dernoncourt, Daniel Preo{\c{t}}iuc-Pietro, and Anastasia Shimorina, editors, \emph{Proceedings of the 2024 Conference on Empirical Methods in Natural Language Processing: Industry Track}, pages 477--485, Miami, Florida, US, November 2024. Association for Computational Linguistics.
\newblock \doi{10.18653/v1/2024.emnlp-industry.36}.
\newblock URL \url{https://aclanthology.org/2024.emnlp-industry.36/}.

\bibitem[Grattafiori et~al.(2024)Grattafiori, Dubey, Jauhri, Pandey, Kadian, Al-Dahle, Letman, Mathur, Schelten, Vaughan, et~al.]{grattafiori2024llama}
Aaron Grattafiori, Abhimanyu Dubey, Abhinav Jauhri, Abhinav Pandey, Abhishek Kadian, Ahmad Al-Dahle, Aiesha Letman, Akhil Mathur, Alan Schelten, Alex Vaughan, et~al.
\newblock The llama 3 herd of models.
\newblock \emph{arXiv preprint arXiv:2407.21783}, 2024.

\bibitem[Hendrycks et~al.(2020)Hendrycks, Burns, Basart, Zou, Mazeika, Song, and Steinhardt]{hendrycks2020measuring}
Dan Hendrycks, Collin Burns, Steven Basart, Andy Zou, Mantas Mazeika, Dawn Song, and Jacob Steinhardt.
\newblock Measuring massive multitask language understanding.
\newblock \emph{arXiv preprint arXiv:2009.03300}, 2020.

\bibitem[Hewitt(2021)]{hewitt2021initializing}
John Hewitt.
\newblock Initializing new word embeddings for pretrained language models, 2021.
\newblock URL \url{https:/nlp.stanford.edu/~johnhew//vocab-expansion.html}.

\bibitem[Hinton et~al.(2015)Hinton, Vinyals, and Dean]{hinton2015distilling}
Geoffrey Hinton, Oriol Vinyals, and Jeff Dean.
\newblock Distilling the knowledge in a neural network.
\newblock \emph{arXiv preprint arXiv:1503.02531}, 2015.

\bibitem[jukofyork(2025)]{jukofyork2025transplant}
jukofyork.
\newblock Vocab transplantation tool (github repository), 2025.
\newblock URL \url{https://github.com/jukofyork/transplant-vocab}.

\bibitem[Kantamneni and Tegmark(2025)]{kantamneni2025language}
Subhash Kantamneni and Max Tegmark.
\newblock Language models use trigonometry to do addition.
\newblock \emph{arXiv preprint arXiv:2502.00873}, 2025.

\bibitem[Kulshreshtha et~al.(2020)Kulshreshtha, Redondo~Garcia, and Chang]{kulshreshtha-etal-2020-cross}
Saurabh Kulshreshtha, Jose~Luis Redondo~Garcia, and Ching-Yun Chang.
\newblock Cross-lingual alignment methods for multilingual {BERT}: A comparative study.
\newblock In Trevor Cohn, Yulan He, and Yang Liu, editors, \emph{Findings of the Association for Computational Linguistics: EMNLP 2020}, pages 933--942, Online, November 2020. Association for Computational Linguistics.
\newblock \doi{10.18653/v1/2020.findings-emnlp.83}.
\newblock URL \url{https://aclanthology.org/2020.findings-emnlp.83/}.

\bibitem[Lee et~al.(2025)Lee, Weber, Viégas, and Wattenberg]{lee2025sharedgloballocalgeometry}
Andrew Lee, Melanie Weber, Fernanda Viégas, and Martin Wattenberg.
\newblock Shared global and local geometry of language model embeddings, 2025.
\newblock URL \url{https://arxiv.org/abs/2503.21073}.

\bibitem[Leviathan et~al.(2023)Leviathan, Kalman, and Matias]{leviathan2023fast}
Yaniv Leviathan, Matan Kalman, and Yossi Matias.
\newblock Fast inference from transformers via speculative decoding.
\newblock In \emph{International Conference on Machine Learning}, pages 19274--19286. PMLR, 2023.

\bibitem[Li et~al.(2024)Li, Fang, Smyrnis, Ivgi, Jordan, Gadre, Bansal, Guha, Keh, Arora, Garg, Xin, Muennighoff, Heckel, Mercat, Chen, Gururangan, Wortsman, Albalak, Bitton, Nezhurina, Abbas, Hsieh, Ghosh, Gardner, Kilian, Zhang, Shao, Pratt, Sanyal, Ilharco, Daras, Marathe, Gokaslan, Zhang, Chandu, Nguyen, Vasiljevic, Kakade, Song, Sanghavi, Faghri, Oh, Zettlemoyer, Lo, El-Nouby, Pouransari, Toshev, Wang, Groeneveld, Soldaini, Koh, Jitsev, Kollar, Dimakis, Carmon, Dave, Schmidt, and Shankar]{li2024datacomplm}
Jeffrey Li, Alex Fang, Georgios Smyrnis, Maor Ivgi, Matt Jordan, Samir Gadre, Hritik Bansal, Etash Guha, Sedrick Keh, Kushal Arora, Saurabh Garg, Rui Xin, Niklas Muennighoff, Reinhard Heckel, Jean Mercat, Mayee Chen, Suchin Gururangan, Mitchell Wortsman, Alon Albalak, Yonatan Bitton, Marianna Nezhurina, Amro Abbas, Cheng-Yu Hsieh, Dhruba Ghosh, Josh Gardner, Maciej Kilian, Hanlin Zhang, Rulin Shao, Sarah Pratt, Sunny Sanyal, Gabriel Ilharco, Giannis Daras, Kalyani Marathe, Aaron Gokaslan, Jieyu Zhang, Khyathi Chandu, Thao Nguyen, Igor Vasiljevic, Sham Kakade, Shuran Song, Sujay Sanghavi, Fartash Faghri, Sewoong Oh, Luke Zettlemoyer, Kyle Lo, Alaaeldin El-Nouby, Hadi Pouransari, Alexander Toshev, Stephanie Wang, Dirk Groeneveld, Luca Soldaini, Pang~Wei Koh, Jenia Jitsev, Thomas Kollar, Alexandros~G. Dimakis, Yair Carmon, Achal Dave, Ludwig Schmidt, and Vaishaal Shankar.
\newblock Datacomp-lm: In search of the next generation of training sets for language models.
\newblock \emph{arXiv preprint arXiv:2406.11794}, 2024.

\bibitem[Menon et~al.(2016)Menon, Gargi, and Samili]{menon2016clustering}
Remya R.~K. Menon, S~Gargi, and S~Samili.
\newblock Clustering of words using dictionary-learnt word representations.
\newblock In \emph{2016 International Conference on Advances in Computing, Communications and Informatics (ICACCI)}, pages 1539--1545, 2016.
\newblock \doi{10.1109/ICACCI.2016.7732267}.

\bibitem[Mikolov et~al.(2013)Mikolov, Chen, Corrado, and Dean]{mikolov2013efficient}
Tomas Mikolov, Kai Chen, Greg Corrado, and Jeffrey Dean.
\newblock Efficient estimation of word representations in vector space.
\newblock \emph{arXiv preprint arXiv:1301.3781}, 2013.

\bibitem[Minixhofer et~al.(2022)Minixhofer, Paischer, and Rekabsaz]{minixhofer-etal-2022-wechsel}
Benjamin Minixhofer, Fabian Paischer, and Navid Rekabsaz.
\newblock {WECHSEL}: Effective initialization of subword embeddings for cross-lingual transfer of monolingual language models.
\newblock In \emph{Proceedings of the 2022 Conference of the North American Chapter of the Association for Computational Linguistics: Human Language Technologies}, pages 3992--4006, Seattle, United States, July 2022. Association for Computational Linguistics.
\newblock URL \url{https://aclanthology.org/2022.naacl-main.293}.

\bibitem[Minixhofer et~al.(2024)Minixhofer, Ponti, and Vulić]{minixhofer2024zeroshottokenizertransfer}
Benjamin Minixhofer, Edoardo~Maria Ponti, and Ivan Vulić.
\newblock Zero-shot tokenizer transfer, 2024.
\newblock URL \url{https://arxiv.org/abs/2405.07883}.

\bibitem[Ostendorff and Rehm(2023)]{Ostendorff2023clp}
Malte Ostendorff and Georg Rehm.
\newblock Efficient language model training through cross-lingual and progressive transfer learning, 2023.

\bibitem[Paperno et~al.(2016)Paperno, Kruszewski, Lazaridou, Pham, Bernardi, Pezzelle, Baroni, Boleda, and Fern{\'a}ndez]{paperno2016lambada}
Denis Paperno, Germ{\'a}n Kruszewski, Angeliki Lazaridou, Quan~Ngoc Pham, Raffaella Bernardi, Sandro Pezzelle, Marco Baroni, Gemma Boleda, and Raquel Fern{\'a}ndez.
\newblock The lambada dataset: Word prediction requiring a broad discourse context.
\newblock \emph{arXiv preprint arXiv:1606.06031}, 2016.

\bibitem[Pati et~al.(1993)Pati, Rezaiifar, and Krishnaprasad]{pati1993orthogonal}
Yagyensh~Chandra Pati, Ramin Rezaiifar, and Perinkulam~Sambamurthy Krishnaprasad.
\newblock Orthogonal matching pursuit: Recursive function approximation with applications to wavelet decomposition.
\newblock In \emph{Proceedings of 27th Asilomar conference on signals, systems and computers}, pages 40--44. IEEE, 1993.

\bibitem[Rust et~al.(2021)Rust, Pfeiffer, Vulić, Ruder, and Gurevych]{rust2021good}
Phillip Rust, Jonas Pfeiffer, Ivan Vulić, Sebastian Ruder, and Iryna Gurevych.
\newblock How good is your tokenizer? on the monolingual performance of multilingual language models.
\newblock pages 3118--3135, 01 2021.
\newblock \doi{10.18653/v1/2021.acl-long.243}.

\bibitem[Skianis et~al.(2018)Skianis, Tziortziotis, and Vazirgiannis]{skianis2018orthogonal}
Konstantinos Skianis, Nikolaos Tziortziotis, and Michalis Vazirgiannis.
\newblock Orthogonal matching pursuit for text classification.
\newblock \emph{arXiv preprint arXiv:1807.04715}, 2018.

\bibitem[Tropp and Gilbert(2007)]{tropp2007signal}
Joel~A Tropp and Anna~C Gilbert.
\newblock Signal recovery from random measurements via orthogonal matching pursuit.
\newblock \emph{IEEE Transactions on information theory}, 53\penalty0 (12):\penalty0 4655--4666, 2007.

\bibitem[Yang et~al.(2024)Yang, Yang, Zhang, Hui, Zheng, Yu, Li, Liu, Huang, Wei, et~al.]{yang2024qwen2}
An~Yang, Baosong Yang, Beichen Zhang, Binyuan Hui, Bo~Zheng, Bowen Yu, Chengyuan Li, Dayiheng Liu, Fei Huang, Haoran Wei, et~al.
\newblock Qwen2. 5 technical report.
\newblock \emph{arXiv preprint arXiv:2412.15115}, 2024.

\bibitem[Yang et~al.(2019)Yang, Zhang, Tar, and Baldridge]{yang2019paws}
Yinfei Yang, Yuan Zhang, Chris Tar, and Jason Baldridge.
\newblock Paws-x: A cross-lingual adversarial dataset for paraphrase identification.
\newblock \emph{arXiv preprint arXiv:1908.11828}, 2019.

\end{thebibliography}

\clearpage
\appendix
\section{Appendix}

\subsection{Token Decomposition Examples}

These examples show how tokens are represented as linear combinations of other tokens in the embedding space.

\begin{figure}[htbp]
\centering
\small
\begin{tabular}{p{1.8cm}p{11cm}}
\toprule
\textbf{Token} & \textbf{Sparse Linear Decomposition} \\
\midrule

\zhchar{读者} & 
\begin{minipage}[t]{11cm}
$\approx$ \contribution{` readers'}{0.147} + 
\contribution{` reader'}{0.129} + 
\contribution{`\zhchar{作者}'}{0.126} + 
\contribution{`\kochar{움}'}{0.139} + 
\contribution{`\zhchar{学生}'}{0.111} + 
\contribution{` retail'}{0.112} + 
\contribution{` visually'}{0.107} + 
\contribution{`\kochar{讀}'}{0.149} 
\end{minipage} \\[2ex]

\midrule
\kochar{불구하고} & 
\begin{minipage}[t]{11cm}
$\approx$ \contribution{`\kochar{그러나}'}{0.142} + 
\contribution{`\kochar{인데}'}{0.122} + 
\contribution{`\kochar{통해}'}{0.101} + 
\contribution{`\jachar{らず}'}{0.094} + 
\contribution{`Despite'}{0.090} + 
\contribution{`\jachar{のに}'}{0.106} + 
\contribution{\textcyrillic{` является'}}{0.090} + 
\contribution{`-D'}{-0.073}
\end{minipage} \\[2ex]

\midrule
\zhchar{它是} & 
\begin{minipage}[t]{11cm}
$\approx$ \contribution{`\zhchar{它}'}{0.245} + 
\contribution{`\zhchar{是}'}{0.126} + 
\contribution{`\zhchar{这是}'}{0.154} + 
\contribution{` \{\}:'}{0.111} + 
\contribution{` Exist'}{0.106} + 
\contribution{`metro'}{0.108} + 
\contribution{`cht'}{-0.095} + 
\contribution{`\zhchar{地说}'}{0.091}
\end{minipage} \\[2ex]

\midrule
` várias' & 
\begin{minipage}[t]{11cm}
$\approx$ \contribution{` varias'}{0.208} + 
\contribution{` diversas'}{0.151} + 
\contribution{` muit'}{0.125} + 
\contribution{` varios'}{0.161} + 
\contribution{` uma'}{0.089} + 
\contribution{` various'}{0.090} + 
\contribution{`ários'}{0.097} + 
\contribution{` Brief'}{-0.077}
\end{minipage} \\[2ex]

\midrule
\zhchar{消費者} & 
\begin{minipage}[t]{11cm}
$\approx$ \contribution{` consumers'}{0.122} + 
\contribution{`\zhchar{市場}'}{0.099} + 
\contribution{`\kochar{고객}'}{0.113} + 
\contribution{`\zhchar{企業}'}{0.101} + 
\contribution{` Consumers'}{0.112} + 
\contribution{`\zhchar{篇}'}{-0.072} + 
\contribution{` visitors'}{0.080} + 
\contribution{` portions'}{-0.078}
\end{minipage} \\[2ex]

\midrule
\zhchar{营利} & 
\begin{minipage}[t]{11cm}
$\approx$ \contribution{`-profit'}{0.141} + 
\contribution{`\zhchar{莉}'}{0.124} + 
\contribution{`.ylim'}{0.127} + 
\contribution{` lucrative'}{0.130} + 
\contribution{`.Windows'}{0.097} + 
\contribution{` purpos'}{0.147} + 
\contribution{`\zhchar{营}'}{0.095} + 
\contribution{`-selling'}{0.106}
\end{minipage} \\

\midrule
    \zhchar{专家} & 
\begin{minipage}[t]{11cm}
$\approx$ \contribution{` experts'}{0.166} + 
\contribution{` expert'}{0.163} + 
\contribution{`\zhchar{教师}'}{0.119} + 
\contribution{`\zhchar{主任}'}{0.118} + 
\contribution{`\zhchar{政策}'}{0.095} + 
\contribution{` Scientists'}{0.125} + 
\contribution{`\zhchar{咨询}'}{0.105} + 
\contribution{`\zhchar{资产}'}{0.098}
\end{minipage} \\


\midrule
\jachar{必要があります} &
\begin{minipage}[t]{11cm}
$\approx$ \contribution{`\jachar{できます}'}{0.197} + 
\contribution{`\kochar{있습니다}'}{0.164} + 
\contribution{\textcyrillic{` необходимо'}}{0.114} + 
\contribution{` harus'}{0.092} + 
\contribution{`\jachar{べき}'}{0.126} + 
\contribution{`However'}{0.086} + 
\contribution{` dispositivo'}{0.115} + 
\contribution{` muss'}{0.092}
\end{minipage} \\

\midrule
\zhchar{氟} &
\begin{minipage}[t]{11cm}
$\approx$ \contribution{` fluoride'}{0.123} +
\contribution{`\zhchar{弗}'}{0.150} +
\contribution{` lithium'}{0.141} +
\contribution{`\zhchar{氧}'}{0.112} +
\contribution{`\includegraphics[width=2.5ex]{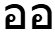}'}{0.139} +
\contribution{` ann'}{-0.103} +
\contribution{` fluor'}{0.134} +
\contribution{` dynam'}{-0.098}
\end{minipage} \\

\midrule
`geführt' &
\begin{minipage}[t]{11cm}
$\approx$ \contribution{` gemacht'}{0.161} +
\contribution{`führt'}{0.202} +
\contribution{`\kochar{한다}'}{0.097} +
\contribution{`geben'}{0.088} +
\contribution{`\jachar{であり}'}{0.076} +
\contribution{`chnitt'}{0.089} +
\contribution{`iliated'}{0.077} +
\contribution{` Sch'}{-0.061}
\end{minipage} \\

\midrule
\bottomrule
\end{tabular}
\caption{Sparse linear decompositions of selected tokens from Qwen 2.5's vocabulary. Each token is decomposed into a weighted sum of $k=8$ basis tokens, with coefficients colored according to magnitude (green for positive, red for negative).}
\label{fig:token-decompositions}
\end{figure}

\subsection{Assets and Software Used}
\label{sec:assets_used}

Table~\ref{tab:assets_software} details the key models, datasets (beyond standard benchmarks cited in text), and software libraries used in this research, along with their sources and licenses, to aid reproducibility. Standard evaluation benchmarks like MMLU, ARC, GSM8K, etc., are cited in the main text and were accessed via the LM Evaluation Harness.

\begin{table}[H]
\centering
\caption{Overview of Key Assets and Software Used.}
\label{tab:assets_software}
\begin{tabular}{p{2.5cm} p{3cm} p{2.5cm} p{3.5cm} p{2cm}}
\toprule
\textbf{Asset Name} & \textbf{Creator / Origin} & \textbf{Version / Identifier} & \textbf{Source / Access} & \textbf{License} \\
\midrule
\multicolumn{5}{l}{\textit{Models}} \\
\midrule
Llama 3.2 1B      & Meta AI~\citep{grattafiori2024llama} & Sept. 2024 release & \url{https://huggingface.co/meta-llama/Llama-3.2-1B} & Llama 3 Community License \\
Qwen 2.5 7B       & Alibaba~\citep{yang2024qwen2} & Sept. 2024 release & \url{https://huggingface.co/Qwen/Qwen2.5-7B} & Apache 2.0 \\
Mistral NeMo 12B  & Mistral AI                  & July 2024 release  & \url{https://huggingface.co/mistralai/Mistral-Nemo-Base-2407} & Apache 2.0 \\
\midrule
\multicolumn{5}{l}{\textit{Software}} \\
\midrule
\verb|transformers|    & Hugging Face                & 4.50.0             & \texttt{pip install transformers==4.50.0} & Apache 2.0 \\
LM Eval Harness   & EleutherAI~\citep{eval-harness} & 0.4.8              & \texttt{pip install lm-eval-harness==0.4.8} & MIT License \\
\verb|mergekit|& Arcee AI~\citep{goddard-etal-2024-arcees} & Hash: \texttt{4da40d2} & \url{https://github.com/arcee-ai/mergekit} & BuSL-1.0 \\ 
\verb|axolotl|         & Axolotl AI    & 0.8.1              & \texttt{pip install axolotl==0.8.1} & Apache 2.0 \\
\bottomrule
\end{tabular}
\end{table}
\end{document}